\documentclass[runningheads]{llncs}
\usepackage{graphicx}
\usepackage{comment}
\usepackage{amsmath,amssymb} 
\usepackage{color}
\usepackage{float}
\usepackage{caption}

\usepackage{epsfig}
\usepackage{graphicx}
\usepackage{amsmath}
\usepackage{amssymb}
\usepackage[normalem]{ulem}
\usepackage[ruled,vlined]{algorithm2e}
\usepackage{bbm}
\usepackage{bm}
\usepackage[tight,normalsize,sf,SF]{subfigure}
\usepackage{seqsplit}

\usepackage{amsmath}

\usepackage[pagebackref=true,breaklinks=true,letterpaper=true,colorlinks,bookmarks=false]{hyperref}

\makeatletter
\DeclareRobustCommand\onedot{\futurelet\@let@token\@onedot}
\def\@onedot{\ifx\@let@token.\else.\null\fi\xspace}
\def\eg{\emph{e.g}\onedot}

\makeatother

\begin{document}
\pagestyle{headings}
\mainmatter
\def\ECCVSubNumber{6965}  

\title{Weakly-Supervised Action Localization with Expectation-Maximization Multi-Instance Learning}

\titlerunning{Weakly-Supervised Action Localization with EM Multi-Instance Learning}
\author{Zhekun Luo$^1$~
Devin Guillory$^1$~  
Baifeng Shi$^2$~  
Wei Ke$^3$~ 
Fang Wan$^4$ \\
Trevor Darrell$^1$~ 
Huijuan Xu$^1$ 
}
\authorrunning{Zhekun Luo et al.}
\institute{$^1$University of California, Berkeley  
$^2$Peking University \\
$^3$Carnegie Mellon University 
$^4$Chinese Academy of Sciences \\
\email{
$^1$\{zhekun\_luo, dguillory, trevordarrell, huijuan\}@eecs.berkeley.edu, \\
$^2$bfshi@pku.edu.cn,
$^3$weik@andrew.cmu.edu, 
$^4$wanfang@ucas.ac.cn}
}

\maketitle
\begin{abstract}
Weakly-supervised action localization  requires training a model to localize the action segments in the video given only video level action label. It can be solved under the Multiple Instance Learning (MIL) framework, where a bag (video) contains multiple instances (action segments). Since only the bag's label is known, the main challenge is assigning which key instances within the bag to trigger the bag's label. Most previous models use attention-based approaches applying attentions to generate the bag's representation from instances, and then train it via the bag's classification. These models, however, implicitly violate the MIL assumption that instances in negative bags should be uniformly negative. In this work, we explicitly model the key instances assignment as a hidden variable and adopt an Expectation-Maximization (EM) framework. We derive two pseudo-label generation schemes to model the E and M process and iteratively optimize the likelihood lower bound. We show that our EM-MIL approach more accurately models both the learning objective and the MIL assumptions. It achieves state-of-the-art performance on two standard benchmarks, THUMOS14 and ActivityNet1.2.


\end{abstract}

\section{Introduction}
\label{sec:intro}
As the growth of video content accelerates, it becomes increasingly necessary to improve video understanding ability with less annotation effort. Since videos can contain a large number of frames, the cost of identifying the exact start and end frames of each action is high (frame-level) in comparison to just labeling what actions the video contains (video-level). Researchers are motivated to explore approaches that do not require per-frame annotations. In this work, we focus on weakly-supervised action localization paradigm, using only video-level action labels to learn activity recognition and localization. This problem can be framed as a special case of the \textbf{Multiple Instance Learning (MIL)} problem \cite{rectanglesMIL}: a bag contains multiple instances; Instances' labels collectively generate the bag's label, and only the bag's label is available during training. In our task, each video represents bag, and the clips of the video represent the instances inside the bag. The key challenge here is to handle \textbf{key instance assignment} during training -- to identify which instances within the bag trigger the bag's label.

\begin{figure}[t]
    \centering
    \includegraphics[width=\linewidth]{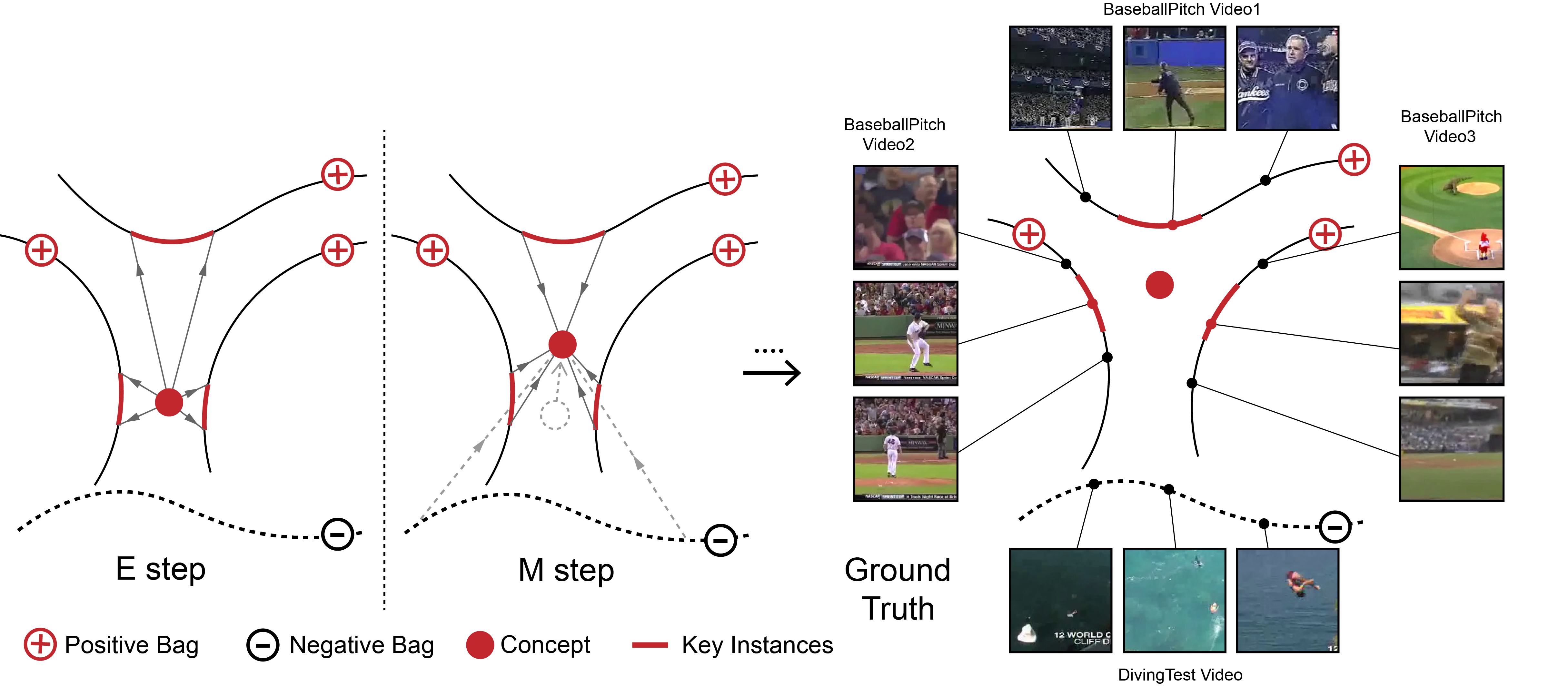}
    \caption{Each curve represents a bag and points on the curve represent instances in the bag. We aim to find a concept point such that each positive bag contains some key instances close to it while all instances in the negative bags are far from it. In E step we use the current concept to pick key instances for each positive bag. In M step we use key instances and negative bags to update the concept.}
    \label{inituition}
\end{figure}

\smallskip
Most previous works used \textbf{attention-based} approaches to model the key instance assignment process. They used attention weights to combine instance-level classification to produce the bag's classification. Models of this form are then trained via standard classification procedures. The learned attention weights imply the contribution of each instance to the bag's label, and thus can be used to localize the positive instances (action clips) \cite{nguyen2018weakly,wang2017untrimmednets}. While promising results have been observed, models of this variety tend to produce incomplete action proposals~\cite{liu2019completeness,yuan2019marginalized}, that only part of the action is detected. This is also a common problem in attention-based weakly-supervised object detection~\cite{Li_2019_ICCV,CMIL}. We argue that this problem is due to a misspecification of the MIL-objective. Attention weights, which indicate key instances' assignment, should be our optimization target. But in an attention-MIL framework, attention is learned as a by-product when conducting classification for bags. As a result, the attention module tends to only pick the most discriminative parts of the action or object to correctly classify a bag, due to the fact that the loss and training signal come from the bag's classification.

\smallskip
Inspired by traditional MIL literature, we adopt a different method to tackle weakly-supervised action localization using the Expectation–Maximization framework. Historically, Expectation–Maximization (EM) or similar iterative estimation processes have been used to solve the MIL problems~\cite{rectanglesMIL,Dooly01multiple-instancelearning,EM_DD} before the deep learning era. Motivated by these works, we explicitly model key instance assignment as a hidden variable and optimize this as our target. Shown in Fig.~\ref{inituition}, we adopt the EM algorithm to solve the interlocking steps of key instance assignment and action concept classification. To formulate our learning objective, we derive two pseudo-label generating schemes to model the E and M process respectively. We show that our alternating update process optimizes a lower bound of the MIL-objective. We also find that previous attention-MIL models implicitly violate the MIL assumptions. They apply attention to negative bags, while the MIL assumption states that instances in negative bags are uniformly negative. We show that our method can better model the data generating procedure of both positive and negative bags. It achieves state-of-the-art performance with a simple architecture, suggesting its potential to be extended to many practical settings. The main contributions of this paper are:

\begin{itemize}
    \item We propose to adapt the Expectation–Maximization MIL framework to weakly supervised action localization task. We derive two novel pseudo-label generating schemes to model the E and M process respectively. \footnote{Code:
    https://github.com/airmachine/EM-MIL-WeaklyActionDetection}

     \item We show that previous attention-MIL models implicitly violate the MIL assumptions, and our method better model the background information. 
    \item Our model is evaluated on two standard benchmarks, THUMOS14 and ActivityNet1.2, and achieves state of the art results. 
    
\end{itemize}

\begin{figure*} [t]
    \centering
    \includegraphics[width=\linewidth]{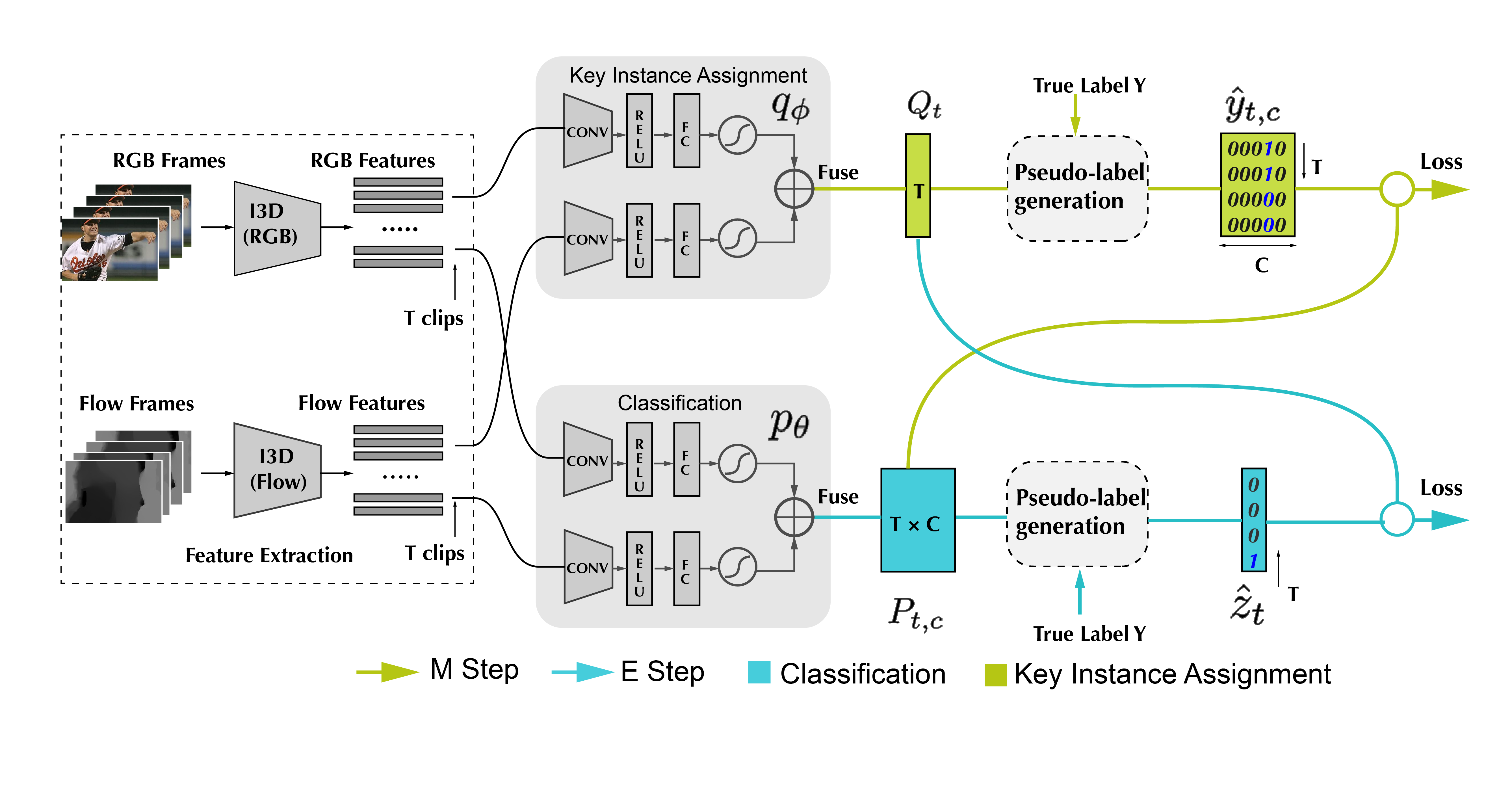}
    \caption{Our EM-MIL model architecture builds on fixed two-stream I3D features, and alternates between updating the key-instance assignment branch \bm{$q_\phi$} (E Step) and the classification branch \bm{$p_\theta$} (M Step). We use the classification score and key instance assignment result to generate pseudo-labels for each other (detailed in Sec.~\ref{E} and Sec.~\ref{M}), and alternate freezing one branch to train the other. 
}
    \label{fig:architecture}
\end{figure*}

\section{Related Work}
\label{sec:related}

\textbf{Weakly-Supervised Action Localization}
Weakly supervised action localization learns to localize activities inside videos when only action class labels are available. UntrimmedNet~\cite{wang2017untrimmednets} first used attention to model the contribution of each clip to a video-level action label. It performs classification separately at each clip, and predicts video's label through a weighted combination of clips' scores. Later the STPN model~\cite{nguyen2018weakly} proposed that instead of combining clips' scores, it uses attention to combine clips' features into a video-level feature vector and conducts classification from there. \cite{ilse2018attention} generalizes a framework for these attention-based approaches and formalizes such combination as a permutation-invariant aggregation function. W-TALC~\cite{paul2018w} proposed a regularization to enforce action periods of the same class must share similar features. It is also noticed that attention-MIL methods tend to produce incomplete localization results. To tackle that, a series of papers~\cite{singh2017hide,su2018cascaded,zeng2019breaking,zhong2018step} took the adversarial erasing idea to improve the detection completeness by hiding the most discriminative parts. \cite{yuan2019marginalized} conducted sub-samplings based on activation to suppress the dominant response of the discriminative action parts. To model complete actions, \cite{liu2019completeness} proposed to use a multi-branch network with each branch handling distinctive action parts. To generate action proposals, they combine per-clip attention and classification scores to form the Temporal Class Activation Sequence (T-CAS \cite{nguyen2018weakly}) and group the high activation clips. Another type of models \cite{shou2018autoloc,liu2019weakly} train a boundary predictor based on pre-trained T-CAS scores to output the action start and end point without grouping. 

\smallskip
Some previous methods in weakly-supervised object or action localization involve iterative refinement, but their  training processes and objectives are different from our Expectation–Maximization method. RefineLoc \cite{alwassel2019refineloc}'s training contains several passes. It uses the result of the $i^{th}$ pass as supervision for the $(i+1)^{th}$ pass and trains a new model from scratch iteratively. \cite{tang2017multiple} uses a similar approach in image objection detection but stacks all passes together. Our approach differs from these in the following ways: Their self-supervision and iterative refinement happen between \textbf{different passes}. In each pass all modules are trained jointly till converge. In comparison, we adopts an EM framework which explicitly models key instance assignment as hidden variables. Our pseudo-labeling and alternating training happen between \textbf{different modules} of the same model. Thus our model requires only one pass. In addition, as discussed in Sec. \ref{Compare}, they handle the attention in negative bags different to us.

\smallskip
\textbf{Traditional Multi-Instance Learning Methods}    
The Multiple Instance Learning problem was first defined by Dietterich et al. \cite{rectanglesMIL}, who proposed the iterated discrimination algorithm. It starts from a point in the feature space and iteratively searches for the smallest box covering at least one point (instance) per positive bag and avoiding all points in negative bags. \cite{DDFramework} sets up the Diverse Density framework. They defined a point in the feature space to be the positive concept. Every positive bag (``diverse") contains at least one instance close to the concept while all instances in the negative bags are far from it (in terms of some distance metric). They modeled the likelihood of a concept using Gaussian Mixture models along with a Noisy-OR probability estimation. \cite{Boosting} then applied AdaBoost to this Noisy-OR model and \cite{ISR}'s ISR model, and derived two MIL loss functions. \cite{Dooly01multiple-instancelearning} adapted the K-nearest neighbors method to the Diverse Density framework. Later \cite{EM_DD} proposed the EM-DD algorithm, combing Expectation Maximization process and the Diverse Density metric. These early works did not involve neural networks and were not applied over the high-dimensional task of action localization. Many of them involve modeling key instances assignment as hidden variable and use iterative optimization. They also differ from the predominant attention-MIL paradigm in how they treat negative instances. We view these distinctions as motivation to explore our approach.

\section{Method}

Multiple Instance Learning (MIL) is a supervised learning problem where instead of one instance $X$ being matched to one label $y$, a bag or set of multiple instances $[X_1, X_2,X_3, ...]$ are matched to single label $y$. In the binary MIL setting, a bag's label is positive if at least one instance in the bag is positive. Therefore a bag is negative only if all instances in the bag are negative.

\smallskip
In our task, following the best practice of previous works~\cite{nguyen2018weakly,paul2018w,wang2017untrimmednets}, we divide a long video into multiple 15-frame clips. Then a video corresponds to a bag (bag-level video label is given), and the clips of the video represent the instances inside the bag (instance-level clip labels are missing). Each video (bag) contains $T$ video clips (instances), denoted by $\mathbf{X} = \{\mathbf{x}_t\}_{t = 1}^{T}$, where $\mathbf{x}_t \in \mathbb{R}^d$ is the feature of clip $t$. 
We represent the video's action label in one hot way, where $y_c = 1$ if the video contains clips of action $c$, otherwise $y_c = 0$, $c \in \{1, 2, \cdots, C\}$ (each video can contain multiple action classes). In the MIL setting, label of each video is determined by the labels of clips it contains. To be specific, we assign a binary variable $z_t \in \{0, 1\}$ to each clip $t$, denoting whether clip $t$ is responsible for the generation of video-level label. $\bm{z} = \{z_t\}_{t = 1}^{T}$ models the assignment of \textbf{key instances’ scope}. Video-level label is generated with probability:
\begin{equation}
    p_\theta(y_c = 1 | \mathbf{X}, \bm{z}) = \sigma_{t \in \{1, \cdots, T\}} \{~ p_\theta(y_{c, t} = 1 | \mathbf{x}_t) \cdot [z_t = 1]~\}
\end{equation}
where $[z_t = 1]$ is the indicator function for assignment. $p_\theta(y_{c, t} = 1 | \mathbf{x}_t)$ is the probability (parameterized by $\theta$) that clip $t$ belongs to class $c$. The closer clip $t$ is to the concept, the higher $p_\theta(y_{c, t} = 1 | \mathbf{x}_t)$ is. $\sigma$ is a permutation-invariant operator, \eg maximum~\cite{zhang2002dd} or mean operator~\cite{ilse2018attention}.

\smallskip
In our temporal action localization problem, we propose to first estimate the probability of $z_t = 1$ with an estimator $q_\phi(z_t = 1 | \mathbf{x}_t)$ parameterized by $\phi$, and then choose the clips with high estimated likelihood as our action segments. Since $\{z_t\}$ are latent variables with no ground truth, we optimize $q_\phi$ through maximization of the variational lower bound:
\begin{equation}
\label{lower_bound}
\begin{split}
    \log p_\theta (y_c | \mathbf{X}) &= KL(q_\phi(\bm{z} | \mathbf{X}) \ || \ p_\theta(\bm{z} | \mathbf{X}, y_c)) + \int q_\phi(\bm{z} | \mathbf{X}) \log \frac{p_\theta(\bm{z}, y_c | \mathbf{X})}{q_\phi(\bm{z} | \mathbf{X})} d \bm{z} \\
    & \geq \int q_\phi(\bm{z} | \mathbf{X}) \log p_\theta(\bm{z}, y_c | \mathbf{X}) d \bm{z} + H(q_\phi(\bm{z} | \mathbf{X})),
\end{split}
\end{equation}

where $H(q_\phi(\bm{z} | \mathbf{X}))$ is entropy of $q_\phi$. By maximizing the lower bound, we are actually optimizing the likelihood of $y_c$ given $\mathbf{X}$. In this work, we adopt the Expectation-Maximization (EM) algorithm, and optimize the lower bound by updating $\theta$ and $\phi$ alternately. To be specific, we first update $\phi$ by minimizing $KL(q_\phi(\bm{z} | \mathbf{X}) \ || \ p_\theta(\bm{z} | \mathbf{X}, y_c))$ and tighten the lower bound in E step, and update $\theta$ through maximization of the lower bound in M step. In the following subsections, we will first get into details of updating $\theta$ and $\phi$ in E step and M step separately, and then sum up the whole algorithm.

\subsection{E Step}\label{E}
In E step, we update $\phi$ by minimizing $KL(q_\phi(\bm{z} | \mathbf{X}) \ || \ p_\theta(\bm{z} | \mathbf{X}, y_c))$ and tighten the lower bound in Eq.~\ref{lower_bound}. As in previous works~\cite{nguyen2018weakly,nguyen2019weakly}, we approximate $q_\phi(\bm{z} | \mathbf{X})$ with $\prod_t q_\phi(z_t | \mathbf{x}_t)$ assuming the independence between different clips, where $q_\phi(z_t | \mathbf{x}_t)$ is estimated by neural network with parameter $\phi$ on each clip. Thus we only have to minimize $KL(q_\phi(z_t | \mathbf{x}_t) \ || \ p_\theta(z_t | \mathbf{x}_t, y_c))$ for each clip $t$. Following the literature, we assume that the posterior $p_\theta(z_t | \mathbf{x}_t, y_c)$ is proportional to the classification score $p_\theta(y_c | \mathbf{x}_t)$. Then we propose to update $q_\phi$ with pseudo label generated from classification score. Specifically, dynamic thresholds are calculated based on the instance classification scores to generate pseudo-labels for $q_\phi$. If an instance has a classification score over the threshold for any ground truth class within the video, the instance is treated as a positive example; otherwise, it is treated as a negative example. The pseudo label is formulated as follows:

\begin{equation}
\begin{small}
    \hat{z}_{t} =\left\{\begin{array}{ll}
    1, & \text{if } \sum_{c=1}^{C}  \mathbbm{1}(P_{t,c} > \overline{P}_{1:T,c} ~\land~  y_{c} = 1 ) > 0 \\
    0, & \textrm{otherwise}
    \end{array}\right.
\label{eq:a}
\end{small}
\end{equation}
where $P_{t,c} = p_\theta(y_c | \mathbf{x}_t)$ and $\overline{P}_{1:T,c}$ is the mean of $P_{t,c}$ over temporal axis.  Then we update $q_\phi$ using binary cross entropy (BCE) loss and the updating process is illustrated in Fig.~\ref{fig:att_labels}.
\begin{equation}
    \mathcal{L}(q_\phi) = - \hat{z}_t \log q_\phi(z_t | \mathbf{x}_t) - (1 - \hat{z}_t) \log (1 - q_\phi(z_t | \mathbf{x}_t)).
\end{equation}

\subsection{M Step}\label{M}

In M step, we update $p_\theta$ through optimization of the lower bound in Eq.~\ref{lower_bound}. Since $H(q_\phi(\bm{z} | \mathbf{X}))$ is constant wrt $\theta$, we only optimize $\int q_\phi(\bm{z} | \mathbf{X}) \log p_\theta(\bm{z}, y_c | \mathbf{X}) d \bm{z}$, which is equivalent to optimize the classification performance given key instance assignment $q_\phi(\bm{z} | \mathbf{X})$. To this end, we use the class-agnostic key-instance assigning module $q_\phi$ and the ground truth video-level labels to generate a $T\times{C}$ pseudo-label map which discriminates between foreground and background clips within the same video. Similarly, our pseudo-label generation procedure calculates a dynamic threshold based on the distribution of instance-assignment scores for each video clip. It assigns positive classifications for all instances whose scores are higher than the threshold, and negative classifications for all instances whose scores are below or instances in negative bags. The pseudo label is given by:
\begin{equation}
\begin{small}
    \hat{y}_{t,c} =\left\{ \begin{array}{ll}
    1, & \text{if } y_{c} = 1 \text{ and } Q_{t} > \overline{Q}_{1:T} + \gamma \cdot (\max(Q_t) - \min(Q_t))\\ 
    0, & \textrm{otherwise}
    \end{array}\right.,
\label{eq:c}
\end{small}
\end{equation}  

where $Q_t = q_\phi(z_t | \mathbf{x}_t)$ and $\overline{Q}_{1:T}$ is the mean of $Q_t$ over temporal axis. The threshold hyper-parameter $\gamma$ implies a distribution priori on how similar the same action exhibits across several videos. Then we update $p_\theta$ with BCE loss and the updating process is illustrated in Fig.~\ref{fig:cls_labels}.
\begin{equation}
    \mathcal{L}(p_\theta) = -\hat{y}_{t,c} \log p_\theta(y_c | x_t) - (1 - \hat{y}_{t,c}) \log (1 - p_\theta(y_c | x_t)).
\end{equation}

\subsection{Overall Algorithm}

We summarize our EM-style algorithm in Alg.~\ref{algo}. We update the key-instance assigning module $q_\phi$ and classification module $p_\theta$ alternately. In E step we freeze the classification $p_\theta$ and update $q_\phi$ using pseudo labels from $p_\theta$. In M step we optimize classification based on $q_\phi$. Two steps are processed alternately to maximize the likelihood $\log p_\theta (y_c | \mathbf{X})$, and meanwhile optimize the localization results. 
\smallskip

\begin{algorithm}[H]
\label{algo}
\SetAlgoLined
 \textbf{Initialization}: learning rate $\beta$, classification threshold $\gamma$\\ 
 classifier parameters $\theta$, \text{attention parameters} $\phi$\ \\
 \While{$\theta, \phi$ \text{has not converged}}{
     $\# E step$ \\
    \For{$(\mathbf{X}, y_c) \  \text{in train set}$}{
     $P_{t, c} \protect\leftarrow p_\theta(y_c | \mathbf{x}_t)$ \;
     $\phi \protect\leftarrow \ \phi - \beta \cdot \nabla_\phi \mathcal{L}(q_\phi)$ \;
  }
    $\# M step$\\    
  \For{$(\mathbf{X}, y_c) \  \text{in train set}$}{
     $Q_t \protect\leftarrow q_\phi(z_t | \mathbf{x}_t)$ \;
     $   \theta \protect\leftarrow \theta - \beta \cdot \nabla_\theta \mathcal{L}(p_\theta)$ \;
     
  } 
 }
 \caption{EM-MIL Weakly-Supervised Activity Localization}
\end{algorithm}

\subsection{Comparison with Previous Methods}\label{Compare}
After careful examination of Eq.~\ref{eq:a} and Eq.~\ref{eq:c}, we find that our pseudo-labeling process $Q_t$ and $\hat{y}_{t,c}$ can also be interpreted as a special kind of attention. Denote loss function by $\mathcal{L}$, then in Eq.~\ref{eq:c}, the loss is calculated as
\begin{equation}
\begin{small}
    \mathcal{L}\left[~p_\theta(\mathbf{y} | \mathbf{x}),~ \mathcal{F}(\mathbf{Q}, \mathbf{y})~ \right]
\label{ours_update}
\end{small}
\end{equation}

\smallskip
$\mathcal{F}$ is the pseudo label generation function in Eq.~\ref{eq:c}, $\mathbf{Q},\mathbf{y},\mathbf{x}$ is the compact expression of $Q_t, y_c, x_t$. On the other hand, if we denote attention and classification score as $\mathbf{a}, \mathbf{c}$, the loss for a typical attention-based model like \cite{wang2017untrimmednets} is:
\begin{equation}
\begin{small}
\mathcal{L}\left[~\bm{\sigma}(\mathbf{c} \odot \mathbf{a}), ~\mathbf{y} ~\right]
\label{their_update}
\end{small}
\end{equation}  

\begin{figure*} [t]
    \centering
    \includegraphics[width=\linewidth]{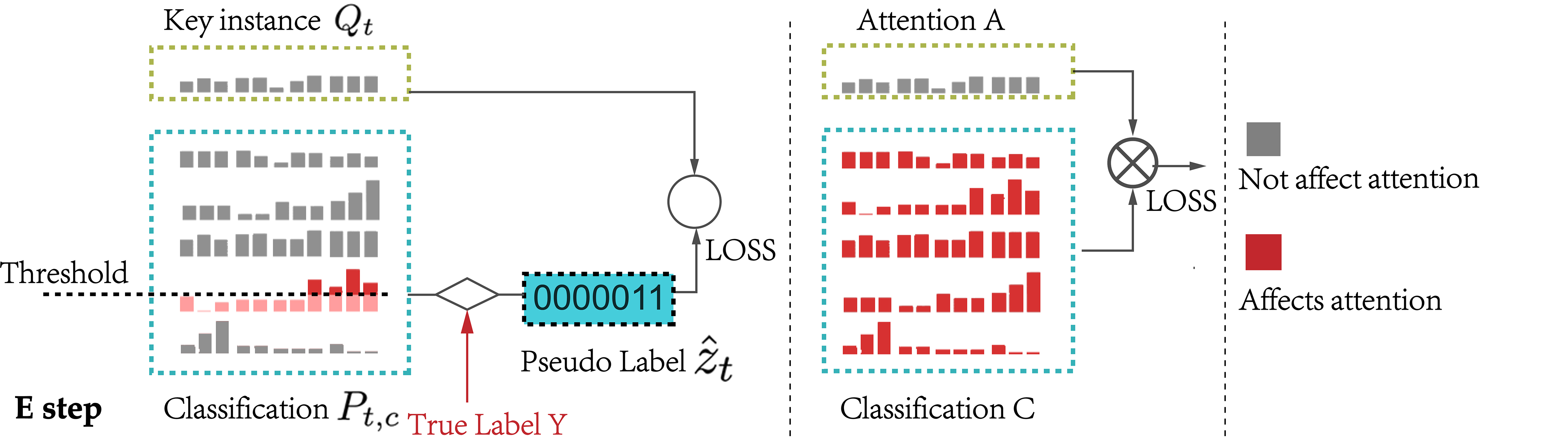}
    \caption{In our EM-MIL model only the foreground classification score $P_{t,c}$ affects the key instance pseudo label $\hat{z}_{t}$ (left), while in previous models all-class classification scores contribute to the attention weights (right).}
    \label{fig:att_labels}
    
    \includegraphics[width=\linewidth]{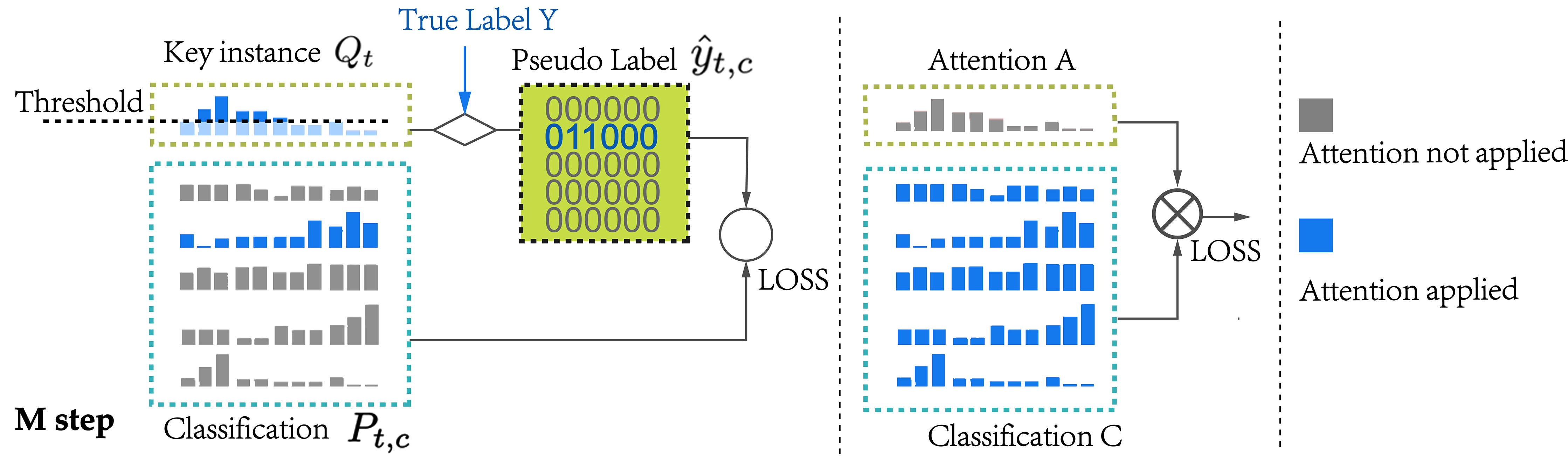}
    \caption{Our EM-MIL model (left) uses key instance assignment $Q_{t}$ to generate pseudo classification labels $\hat{y}_{t,c}$ only for the foreground classes, while in previous models such as UntrimmedNet (right) attentions are applied to all classes.}
    \label{fig:cls_labels}
    
\end{figure*}

\smallskip
Here $\bm{\sigma}$ is the aggregation operator~\cite{ilse2018attention}, such as $reduce\_sum$ or $reduce\_max$. Comparing Eq.~\ref{ours_update} to Eq.~\ref{their_update}, it is easy to see that they can be matched. $p_\theta(\mathbf{y} | \mathbf{x})$ is classification score ($\mathbf{c}$), and $\mathbf{Q}$ can be seen as special attention (corresponds to $\mathbf{a}$). In M step it attends to the key instance it estimates. But compared to previous attention-MIL methods, Eq.~\ref{eq:a} shows that this ``attention" only happens in positive bags. We believe it better aligns with the MIL assumption, which says that all instances in negative bags are uniformly negative. Previous methods that applies attention to negative bags implicitly assumes that some instances are more negative than others. This violates the MIL assumption. The differences between our attention and theirs are illustrated in Fig.~\ref{fig:att_labels} and~\ref{fig:cls_labels}. In addition, in Eq.~\ref{eq:c}, this ``attention" is a threshold-based hard attention. Clips below the threshold are classified as background with high confidence, while clips above the threshold are weighted equally and re-scored in the next iteration. The use of hard pseudo labels allows for the distinct treatment of positive and negative instances that would be more complex to enforce with soft-boundaries. We initialize our training procedure by labeling every clip in a positive bag to be 1 and gradually narrow down the search scope. Such training process maintains high recalls for action clips in each E-M iteration. It prevents attention from focusing on the discriminative parts too quickly, thus increases the proposal completeness.

\smallskip
Another way to compare our methods with previous ones is through the lens of the MIL framework. As discussed in \cite{survey-MIL}, MIL problem has two setting: instance-level vs bag-level. The instance level setting prioritizes classification precision of instance over bag's, and vice versa. Our task aligns with the instance setting as the primary goal is action localization (equivalent to clips' classification). Previous attention-MIL models like \cite{nguyen2018weakly,paul2018w,wang2017untrimmednets} treat instance-localization as the by-product of an accurate bag-level classification system, which align with the bag-level MIL setting. By modeling the problem through an instance-level MIL framework our approach more accurately models the target objective. This change in objective function and optimization procedure allows substantial improvement in performance.

\subsection{Inference}
At test time, we use another branch for video-level classification and use our model for localization as in previous work~\cite{shou2018autoloc}. For classification branch, we used a plain UntrimmedNet \cite{wang2017untrimmednets} with soft attention for the THUMOS14 dataset and the W-TALC \cite{paul2018w} for the ActivityNet1.2 dataset. We run a forward pass with our model to get the localization score $L$ by fusing instance assignment score $Q_t$ and classification score $P_{t,c}$.
\begin{equation}
\label{test}
    L_t = \lambda*Q_t + (1-\lambda)*P_{t,c},
\end{equation}
where $\lambda$ is set to be 0.8 through grid search in THUMOS14 dataset and 0.3 in the ActivityNet1.2 dataset. 
In the Experiment Sec.~\ref{sec:exp:SOA} we analyze the impact of different of $\lambda$. We threshold the $L_t$ score to get prediction $y_{i}'$ for each clip using the same scheme as in Eq.~\ref{eq:c}. Then we group the clips above the threshold to get the temporal start and end point of the action proposal.

           
\section{Experiments}
\label{sec:exp}
In this section, we evaluate our EM-MIL model on two large-scale temporal activity detection datasets: THUMOS14~\cite{THUMOS14} and ActivityNet1.2~\cite{caba2015activitynet}. Sec.~\ref{sec:exp:setup} introduces experimental setup of these datasets, the evaluation metrics and the implementation details. Sec.~\ref{sec:exp:SOA} compares weakly localization results between our proposed model and the state-of-the-art models on both THUMOS14 and ActivityNet1.2 datasets, and visualizes some localization results. Sec.~\ref{sec:exp:ablation} shows the ablation studies for each component of our model on THUMOS14 dataset.

\subsection{Experimental  Setup}
\label{sec:exp:setup}

\noindent{\textbf{Datasets}:}
The THUMOS14~\cite{THUMOS14} activity detection dataset contains over 24 hours of videos from 20 different athletic activities. The train set contains 2765 trimmed videos, while the validation set and the test set contains 200 and 213 untrimmed videos respectively. 
We use the validation set as train data and report weakly-supervised temporal activity localization results on the test set. This dataset is particularly challenging as it consists of very long videos with multiple activity instances of very small duration. Most videos contain multiple activity instances of the same activity class. In addition, some videos contain activity instances from different classes.

\smallskip
The ActivityNet~\cite{caba2015activitynet} dataset consists three versions. We use the ActivityNet1.2 version which contains a total of around 10000 videos including 4819 train videos, 2383 validation videos, and 2480 withheld test videos for challenge purpose. 
We report the weakly-supervised temporal activity localization results on the validation videos. In ActivityNet1.2, around 99\% videos contain activity instances of a single class. Many of the videos have activity instances covering more than half of the duration. Compared to THUMOS14, this is a large-scale dataset, both in terms of the number of activities involved and the amount of videos.

\smallskip
\noindent{\textbf{Evaluation Metric}:}
The weakly-supervised temporal activity localization results are evaluated in terms of mean Average Precision (mAP) with different temporal Intersection over Union (tIoU) thresholds, which is denoted as mAP@$\alpha$ where $\alpha$ is the threshold. Average mAP at 10 evenly distributed tIoU thresholds between 0.5 and 0.95 is also commonly used in the literature.

\smallskip
\noindent{\textbf{Implementation Details}:} 
Video frames are sampled at 12 fps (for THUMOS14) or 25 fps (for ActivityNet1.2). For each frame, we perform the center crop of size $224 \times 224$ after re-scaling the shorter dimension to 256 and construct video clips for every 15 frames. We extract the features of the clips using the publicly released, two-stream I3D model pretrained on Kinetics dataset~\cite{Carreira2017QuoVA}. We use the feature map from $Mixed\_5c$ layer as feature representation. For optical flow stream, TV-L1 flow~\cite{wang2016temporal,zach2007duality} is used as the input.

\smallskip
Our model is implemented in pyTorch and trained using Adam optimizer with initial learning rate 0.0001 for both datasets. For the THUMOS14 dataset, we train the model by alternating E/M step every 10 epochs in the first 30 epochs. Then we raise the learning rate to 4 times larger and decrease the alternating cycle to 1 epoch for another 35 epochs. For ActivityNet1.2 dataset, we use a similar training approach but the alternating cycle is 5 epochs and the learning rate is constant. 
We use our model to generate instance assignment $Q_t$ and classification score $P_{t,c}$ separately for RGB and Flow branch. Then, we fuse the RGB/Flow score by weighted averaging. The threshold hyper-parameter $\gamma$ in Eq.~\ref{eq:c} is set to 0.15 for THUMOS14 dataset and 0 for ActivityNet1.2 dataset. Intuitively, the value of $\gamma$ reflects how similar the same action exhibits across several videos, and should be negatively correlated with the variance of the action's feature distribution. We also explore different $\gamma$ in the range of [0.05, 0.2], mAP@tIoU=0.5 varies between 29.0\% and 30.5\% in THUMOS14 dataset, compared to the previous SOTA 26.8\%~\cite{nguyen2019weakly} using the same training data.

\begin{table*}[!t]
\centering
\caption{Our EM-MIL detection results on THUMOS14 in percentage. mAP at different tIoU thresholds $\alpha$ are reported. The top half shows fully-supervised methods while the bottom half shows weakly-supervised ones including ours. EM-MIL-UNT represents the result using UntrimmedNet's \cite{wang2017untrimmednets} features.}

\begin{tabular}{l| l || c c c c c c c} 
\hline
 ~& ~ & \multicolumn{7}{c}{$\alpha$} \\
 Supervision~~ & Models & \!\!~~0.1~~ & \!\!~~0.2~~ & \!\!~~0.3~~ & \!\!~~0.4~~ & \!\!~~0.5~~ & \!\!~~0.6~~ & \!\!~~0.7~~ \\ \hline

~& \!\! CDC~\cite{shou2017cdc} \!\!\!& \!\!- & \!\!-  & \!\!40.1  & \!\!29.4 & \!\!23.3  & \!\!13.1 & \!\!7.9  \\

~& \!\! R-C3D~\cite{xu2017r} \!\!\!& \!\!54.5 & \!\!51.5  & \!\!44.8  & \!\!35.6 & \!\!28.9  & \!\!- & \!\!-  \\ 
Fully- & \!\! Gao et al.~\cite{gao2017cascaded} \!\!\!& \!\!-& \!\!-  & \!\!50.1   & \!\!41.3 & \!\!31.0  & \!\!19.1 & \!\!9.9  \\ 
Supervised~& \!\! SSN~\cite{zhao2017temporal} \!\!\!& \!\!66.0& \!\!59.4  & \!\!51.9   & \!\!41.0 & \!\!29.8  & \!\!19.6 & \!\!10.7  \\ 

~& \!\! Xu et al.~\cite{xu2019two} \!\!\!& \!\!56.9& \!\!54.7  & \!\!51.2   & \!\!43.0 & \!\!36.1  & \!\!- & \!\!-  \\ 

~& \!\! BSN~\cite{lin2018bsn} \!\!\!& \!\!- & \!\!-  & \!\!53.5  & \!\!45.0 & \!\!36.9  & \!\!28.4 & \!\!20.0  \\ 
\hline

~ & \!\! Hide~\cite{singh2017hide} & \!\!36.4 & \!\! 27.8  & \!\!19.5  & \!\! 12.7  & \!\!6.8  & \!\!- & \!\!-  \\

~ & \!\! UntrimmedNet~\cite{wang2017untrimmednets}& \!\!44.4 & \!\! 37.7  & \!\!28.2  & \!\! 21.1  & \!\!13.7  & \!\!- & \!\!-  \\ 

~& \!\! STPN~\cite{nguyen2018weakly} & \!\!52.0 & \!\!44.7  & \!\!35.5  & \!\!25.8 & \!\!16.9  & \!\!9.9 & \!\!4.3  \\ 
~& \!\! Autoloc~\cite{shou2018autoloc} & \!\!- & \!\!-  & \!\!35.8  & \!\!29.0 & \!\!21.2  & \!\!13.4 & \!\!5.8  \\ 
~& \!\! W-TALC~\cite{paul2018w} & \!\!55.2 & \!\!49.6  & \!\!40.1  & \!\!31.1 & \!\!22.8  & \!\!- & \!\!7.6  \\ 

~ & \!\! RefineLoc-I3D~\cite{alwassel2019refineloc} & \!\!- & \!\! -  & \!\!40.8  & \!\! -  & \!\!23.1  & \!\!- & \!\!5.3  \\

Weakly- & \!\! Liu et al. ~\cite{liu2019weakly} & \!\! - & \!\! -  & \!\!37.0  & \!\! 30.9  & \!\!23.9  & \!\!13.9 & \!\!7.1  \\ 
Supervised& \!\! Yu et al. ~\cite{yu2019temporal} & \!\! - & \!\! -  & \!\!39.5  & \!\! -  & \!\!24.5  & \!\!- & \!\!7.1  \\ 
~& \!\! 3C-Net~\cite{narayan20193c} & \!\!59.1 & \!\!53.5  & \!\!44.2  & \!\!34.1 & \!\!26.6  & \!\!- & \!\!8.1  \\ 
~& \!\! Nguyen et al. ~\cite{nguyen2019weakly} & \!\! \textbf{64.2} & \!\! \textbf{59.5}  & \!\!~\textbf{49.1}  & \!\! \textbf{38.4}  & \!\!27.5  & \!\!17.3 & \!\!8.6  \\ 

~& \textbf{EM-MIL (ours)} & 59.1 & 52.7 & 45.5 & 36.8  &\textbf{30.5}  & \textbf{22.7} & \textbf{16.4} \\ 
~& EM-MIL-UNT (ours)  & 59.0 & 50.4 & 42.7 & 34.5  &27.2 & 18.9 & 10.2 \\ \hline

\end{tabular}

\label{tab:res_thumos14}

\end{table*}

\begin{figure*}[!t]
\centering
\subfigure[THUMOS14]{
\includegraphics[width=0.7\linewidth]{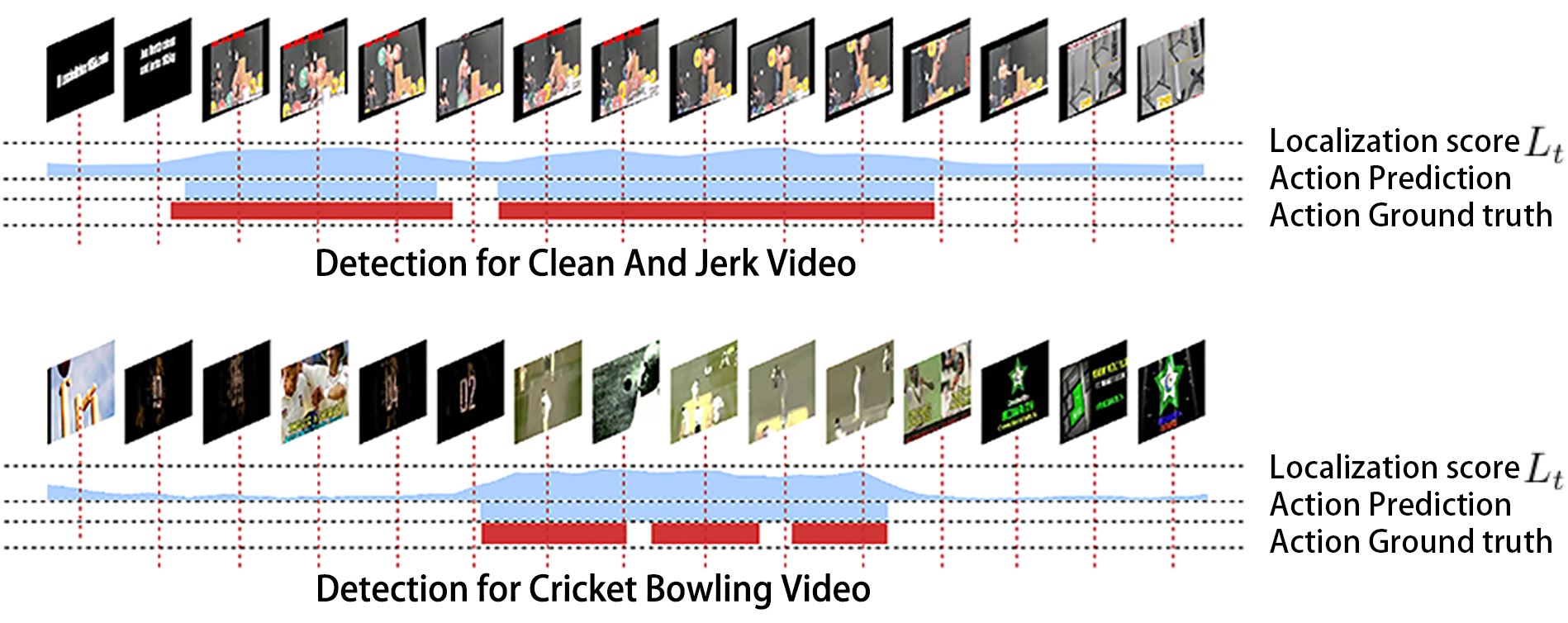}
\label{fig:vis_thumos14}
}
\subfigure[ActivityNet1.2]{
\label{fig:vis_activitynet}
\includegraphics[width=0.7\linewidth]{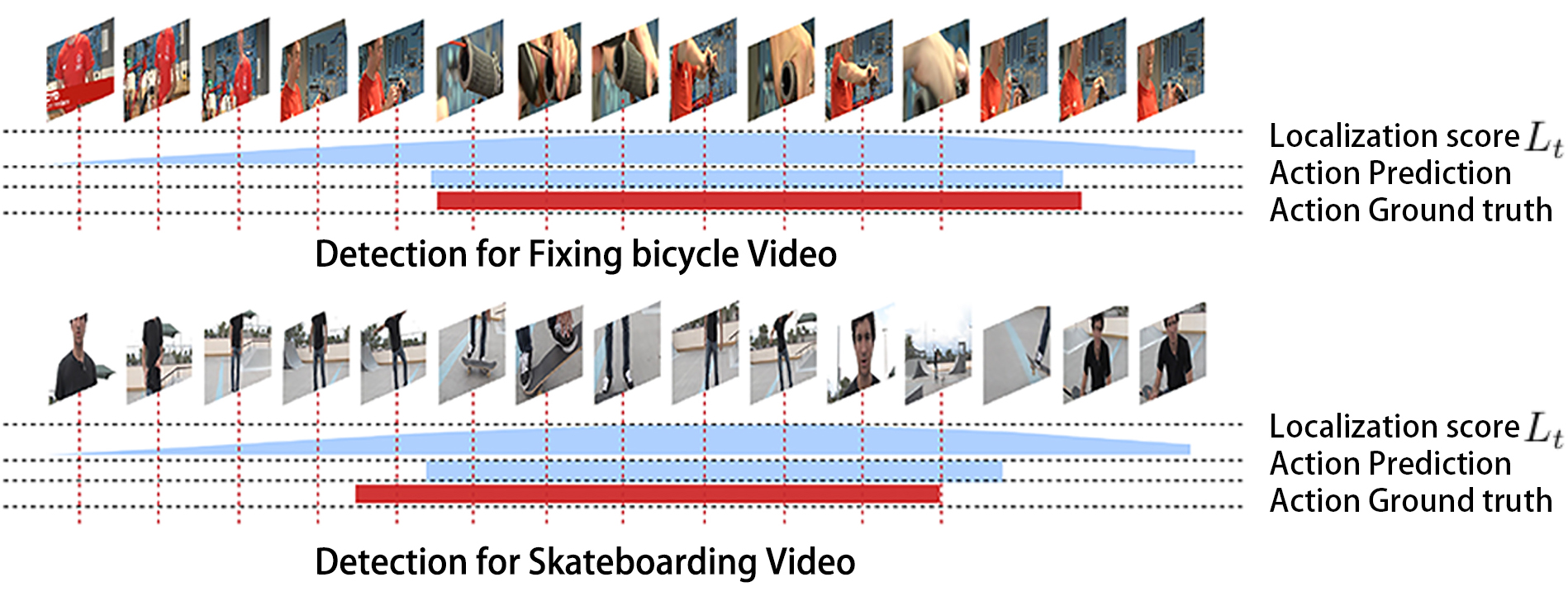}}

\caption{Qualitative visualization. \subref{fig:vis_thumos14} and \subref{fig:vis_activitynet} show results for two videos each on THUMOS14 and ActivityNet1.2, a good prediction example (top) and a bad one (bottom). Ground truth activity segments are marked in red. Localization score distribution $L_t$ and predicted activity segments are in blue.}
\label{fig:qualitative}

\end{figure*}

\subsection{Comparison with State-of-the-art Approaches}
\label{sec:exp:SOA}

\noindent{\textbf{Results on THUMOS14 Dataset}:}
We compare our model's results on the THUMOS14 dataset with state-of-the-art results in Table~\ref{tab:res_thumos14}.
Our model outperforms all the previous published models and achieves a new state-of-the-art result at mAP@0.5, \textbf{30.5\%}. This result is achieved by our simple EM training policy and the pseudo-labeling scheme, without auxiliary losses to regularize the learning process. 
Compared to the best result among the six recent models~\cite{alwassel2019refineloc,narayan20193c,nguyen2018weakly,nguyen2019weakly,paul2018w,yu2019temporal} using the same two-stream I3D feature extraction backbone as our model, we get 3\% significant improvement at mAP@0.5. 
We also tried using UntrimmedNet's feature on our model (denoted as EM-MIL-UNT in Table~\ref{tab:res_thumos14}), and got a mAP@0.5 of 27.2\% which still improves significantly over previous models (e.g. \cite{liu2019weakly,shou2018autoloc,wang2017untrimmednets}) using the same feature backbone.
Our model also shows more significant improvement at high threshold metrics  tIoU=0.6 and tIoU=0.7, which implies that our action proposals are more complete. On the other hand, our performance is slightly worse in the low IoU metrics.

\smallskip
Several examples' qualitative results are shown in Fig.~\ref{fig:vis_thumos14}.
For each example, we show the video, intermediate score map $L_t$ from our model, final activity detection result and ground truth temporal segment annotation.
In the first example of $Clean~and~Jerk$, we localize the activity correctly with almost 100\% overlap. We also show one bad prediction from our model in the second example, where our model overestimates the $Cricket~Bowling$ activity duration by 20\%, as an effect of the interactive shrinkage training process which first labels every instance positive. Our model greatly resolves the incompleteness problem for activity detection in videos containing multiple action segments, while in some cases it might also bring in additional false positives. In addition, our model is also highly time efficient: in THUMOS14 our model trains for 65 epochs, taking 64.7s on two TITAN RTX GPUs. We have run the released code for AutoLoc \cite{shou2018autoloc} and W-TALC \cite{paul2018w} on the same machine with their recommended training procedures. Their training times are 44.5s and 6051.2s, respectively. All experiments used pre-computed features and \cite{shou2018autoloc}'s training required additional pretrained CAS scores.

\smallskip
\noindent{\textbf{Results on ActivityNet1.2 Dataset}:}
We compare our model's results on the ActivityNet1.2 dataset with previous results in Table~\ref{table:res:activitynet}. Our model outperforms previously published models in mAP@0.5 and gets the value of \textbf{37.4\%}. Despite the state-of-the-art result in mAP@0.5, our model performs worse in high tIoU metrics, which is the opposite to what we observed on THUMOS14 dataset. 
We further investigate the reason for different result trends on both datasets. Videos in the THUMOS14 dataset contains multiple action segments, each segment with relatively short duration.
It has high localization requirement where our model outperforms pervious ones at high tIoU.
Unlike THUMOS14, most videos ($>$ 99\%) in the ActivityNet1.2 dataset have only one action class, and most of these videos have only a few activity segments which compose a big portion of the whole video duration.
Thus videos in ActivityNet1.2 dataset can be regarded as trimmed actions in certain extent. We speculate that the action localization performance in the ActivityNet1.2 dataset depends more on the classification module, which might be the bottleneck for our model. This speculation also correlates with the different $\lambda$ values in Eq.~\ref{test} when calculating localization score on THUMOS14 and ActivityNet1.2 datasets. 
According to our model's assumption, key instance assignment score $Q_t$ implies the action clips and higher weight for this part facilitates the localization. On THUMOS14, the weight $\lambda$ for the key instance assignment score $Q_t$ is set to be a high value 0.8.
But for ActivityNet1.2, the classification score $P_{t,c}$ has a higher weight (0.7), implying that the model mostly relies on classification to succeed on this dataset. 
For further illustration, we also visualize some good and bad detection results from ActivityNet1.2 dataset in Fig.~\ref{fig:vis_activitynet}.

\begin{table}[!t]
\centering
\caption{Detection results on ActivityNet1.2 in terms of mAP@\{0.5, 0.7, 0.9\} and average mAP at tIoU thresholds $\alpha \in (0.5,0.95)$ with step 0.05 (in percentage). It shows both fully-supervised method and weakly-supervised ones.
}

\small
\begin{tabular}{l| l || c c c| c} 
\hline
 ~& ~& ~& $\alpha$&~ & ~ \\
Supervision~~ & Models & ~~0.5~~  & ~~0.7~~  & ~~0.9~~ & ~~avg. mAP~~ \\ \hline
Fully-Supervised & SSN~\cite{zhao2017temporal} & 41.3 & 30.4  & 13.2  & 26.6\\  \hline

~ & UntrimmedNet~\cite{wang2017untrimmednets} & 7.4 & 3.9  & 1.2  & 3.6\\
~ & Autoloc~\cite{shou2018autoloc} & 27.3 & 17.5  & 6.8  & 16.0\\
~ & W-TALC~\cite{paul2018w} & 37.0 & 14.6  & 4.2 & 18.0\\
Weakly-Supervised~~& 3C-Net~\cite{narayan20193c}  & 37.2 & \textbf{23.7}  & \textbf{9.2}  & \textbf{21.7}\\
~ &  Liu et al. ~\cite{liu2019weakly}  & 37.1& 23.4  & 9.2  & 21.6\\
~ & TSM ~\cite{yu2019temporal}  & 28.3& 18.9  & 7.5  & 17.1\\
~& \textbf{EM-MIL (ours)~} & \textbf{37.4} & 23.1  & 2.0  & 20.3\\ 
\hline

\end{tabular}

\label{table:res:activitynet}

\end{table}

\subsection{Ablation Studies}
\label{sec:exp:ablation}

We ablate our pseudo label generation scheme and Expectation-Maximization alternating training method on THUMOS14 dataset with mAP@0.5 in Table~\ref{table:Ablation:THUMOS14}.

\smallskip
\noindent{\textbf{Ablation on the Pseudo Labeling}:}
We first ablate on the pseudo labeling scheme for $\hat{z}_{t}$ and $\hat{y}_{t,c}$, and include the results in Table~\ref{table:Ablation:THUMOS14}. We switch our learning to be supervised by an attention-MIL loss based on softmax function, similar to~\cite{nguyen2018weakly,wang2017untrimmednets}. In the E step, classification scores of all classes contribute collectively to the attention weights. In the M step, attention weights are applied equally to both positive and negative videos without paying special attention to the bag's label. Compared to the ``Alternating model" doing alternating training but with a plain attention, ``Full Model" improves mAP@0.5 from 24.5\% to 30.5\%. This indicates the usefulness of the proposed pseudo labeling strategy. It models the key instance assignment explicitly and aligns with the MIL assumption better.

\smallskip
\noindent{\textbf{Ablation on the EM Alternating Training Technique}:}
We also evaluate the effectiveness of Expectation-Maximization alternating training compared to joint optimization. The EM training method iteratively estimates the key instance assignment, then maximizes the video classification accuracy, and achieves better activity detection performance. ``Full Model" improves mAP@0.5 from 26.8\% to 30.5\% compared to ``Pseudo labeling" model with joint optimization. The same training process can be potentially applied on other MIL based models for weakly-supervised object detection task to improve accuracy as well.

~\\
\begin{table}[t]
\centering
\caption{Ablation results for the pseudo labeling and EM alternating training on THUMOS14 dataset in terms of mAP@0.5 (\%). }

\small
\begin{tabular}{l| l l || c }  
\hline
\!\!Ablation Models  & ~~Pseudo Label & ~~~  \!\!Alternating Training~~ & \!\! ~mAP@0.5 \\ \hline
\!\!Alternating model~ &  & ~~~~~~~~~~~~\checkmark  & 24.5  \\ 
\!\!Pseudo labeling model~ &~~\checkmark &  & 26.8  \\ 
\!\!Full Model ~&~~\checkmark & ~~~~~~~~~~~~\checkmark & 30.5  \\ \hline
\end{tabular}
\label{table:Ablation:THUMOS14}

\end{table}

\section{Conclusion}
\label{sec:conclude}
We propose a EM-MIL framework with pseudo labeling and alternating training for weakly-supervised action detection in video.
Our EM-MIL framework is motivated by traditional MIL literature which is under-explored in deep learning settings. By allowing us to explicitly model latent variables, this framework improves our control over the learning objective of the instance-level MIL, which leads to state of the art performance. 
While this work uses a relatively simple pseudo-labeling scheme to implement the EM method, more sophisticated EM methods can be designed, e.g. explicitly parameterize the latent distribution for instances and directly optimize the instance likelihood in E and M steps.
Incorporating the video's temporal structure is also a promising direction for further performance improvement.

\section*{Acknowledgement}

Prof. Darrell’s group was supported in part by DoD, BAIR and BDD.

\clearpage
\bibliographystyle{splncs04}
\bibliography{egbib}
\end{document}